\pdfoutput=1

\documentclass[11pt]{article}

\usepackage[]{naacl2021}

\usepackage{times}
\usepackage{latexsym}
\usepackage{graphics}
\usepackage{graphicx}
\usepackage{float}
\usepackage{booktabs}
\usepackage{multirow}
\usepackage[T1]{fontenc}

\usepackage[utf8]{inputenc}

\usepackage{amssymb}
\usepackage{pifont}
\newcommand{\cmark}{\ding{51}}%
\newcommand{\xmark}{\ding{55}}%
\usepackage{microtype}

\title{Leveraging Slot Descriptions for Zero-Shot Cross-Domain Dialogue State Tracking}

\author{Zhaojiang Lin$^1$\thanks{\quad Work done during internship at Facebook}$^*$, Bing Liu$^2$, Seungwhan Moon$^2$, Paul Crook$^2$, Zhenpeng Zhou$^2$,  \\ \textbf{Zhiguang Wang$^2$, Zhou Yu$^3$, Andrea Madotto$^1$$^*$, Eunjoon Cho$^2$, Rajen Subba$^2$}\\
  $^1$The Hong Kong University of Science and Technology \\
  $^2$Facebook \\
  $^3$Columbia University \\
  \texttt{zlinao@ust.hk}, \texttt{bingl@fb.com}\\

  }

\begin{document}
\maketitle
\begin{abstract}



Zero-shot cross-domain dialogue state tracking (DST) enables us to handle task-oriented dialogue in unseen domains without the expense of collecting in-domain data. In this paper, we propose a slot description enhanced generative approach for zero-shot cross-domain DST. Specifically, our model first encodes dialogue context and slots with a pre-trained self-attentive encoder, and generates slot values in an auto-regressive manner. In addition, we incorporate \textit{Slot Type Informed Descriptions} that capture the shared information across slots to facilitate cross-domain knowledge transfer. Experimental results on the MultiWOZ dataset show that our proposed method significantly improves existing state-of-the-art results in the zero-shot cross-domain setting.



\end{abstract}



\section{Introduction}
Task-oriented dialogue systems are designed to assist users in performing daily activities, such as restaurant booking, travel planning, and online shopping. These virtual assistants provide natural language interfaces to services and online APIs~\cite{rastogi2020towards}. Based on users' needs, these systems frequently require support for new domains. However, the current state-of-the-art systems require a substantial amount of in-domain data to properly model a new domain. The data-collection process is both expensive and time-consuming, and thus it is very important to study methods that can build robust and scalable dialogue systems using little to no in-domain data. 

The dialogue state tracking (DST) is an essential component of task-oriented dialogue systems that tracks users' requirements over multi-turn conversations. A popular formulation of the dialogue state is in the form of a list of slot-value pairs. In DST, tracking unseen slots in a new domain, a.k.a. zero-shot domain adaptation, is a significant challenge, since the model has never seen in-domain training samples. There are two main lines of work to tackle this problem. The first proposes domain transferable models using copy mechanisms or ontology graph information~\cite{wu2019transferable,zhou2019multi}. A limitation of such models is that they may not fully leverage pre-trained language models due to the specialized model architecture. The second line of work uses slot-descriptions as input to the model to facilitate the slot understanding~\cite{rastogi2020towards}. However, the provided slot descriptions are collected by crowd sourced human annotators and might be inconsistent among different domains. In general, the optimal approach for constructing slot descriptions in zero-shot settings remains unexplored.

\begin{figure}[t]
    \centering
    \includegraphics[width=0.8\linewidth]{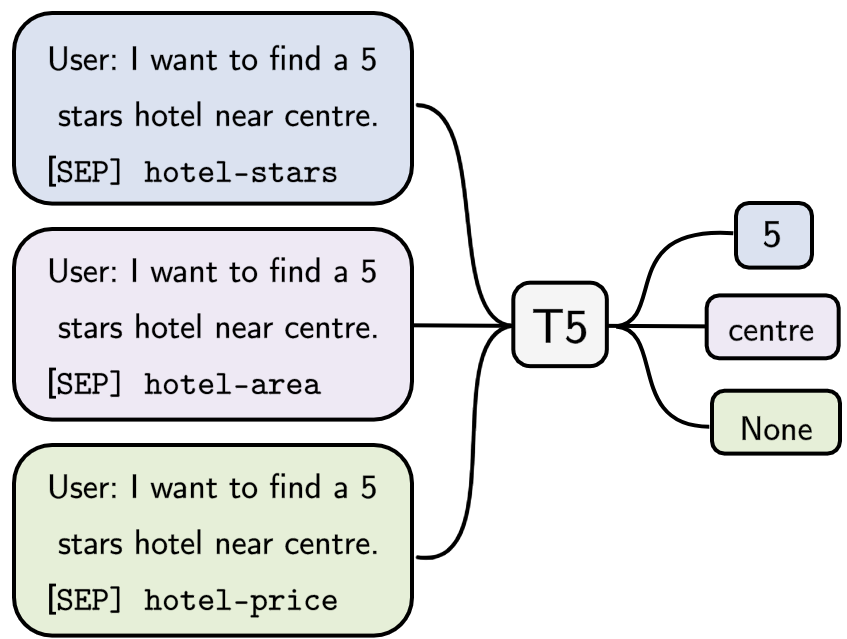}
    \caption{High-level description of the T5DST. The model takes dialogue history and slot name as input, and generates the value.}
    \label{fig:diagram}
\end{figure}

In this work, we tackle the challenge of zero-shot cross-domain DST via leveraging large scale pre-trained sequence-to-sequence (seq2seq) models and with effective encoding of slot descriptions. We first introduce a generative DST model called T5DST, which models the relation of a slot and its dialogue context with a self-attentive encoder, and generates the slot value with a decoder in an autoregressive manner. This simple design allows us to effectively incorporate a pre-trained seq2seq model (e.g., T5~\cite{raffel2020exploring}) without any task-specific modification. To further enhance the model's cross-domain transferability, we propose \textit{Slot Type Informed Descriptions} that capture the shared information of different slots. Experimental results on the MultiWOZ benchmark~\cite{budzianowski2018multiwoz} suggest that 1) our model achieves significantly higher joint goal accuracy compared to existing results in zero-shot cross domain DST; 2) models using the proposed slot description formulation substantially outperform those using other slot description variants. Our contributions are summarized as the following:

\begin{itemize}
    \item We propose a simple yet novel generative DST model based on T5 that significantly improves existing zero-shot cross-domain DST results;
    \item We investigate the effectiveness of different slot description formulations. To the best of our knowledge, this is the first work that comprehensively studies the effectiveness of slot descriptions in zero-shot cross-domain DST.
\end{itemize}

\begin{figure}[t]
    \centering
    \includegraphics[width=0.8\linewidth]{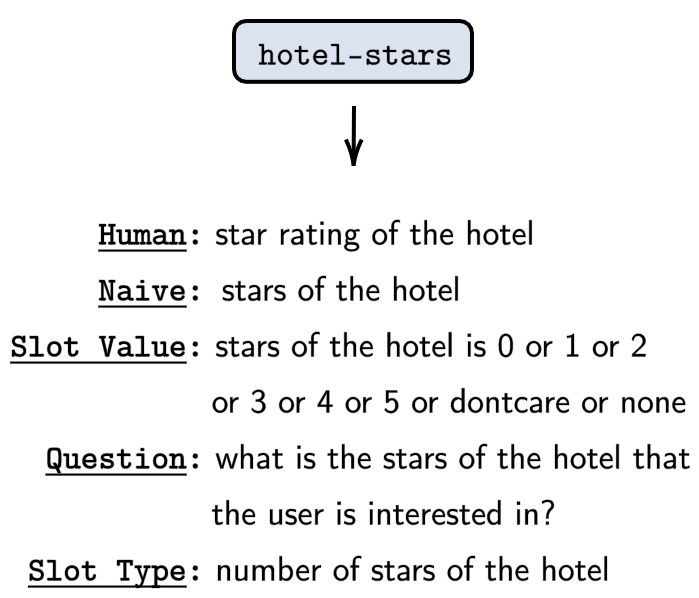}
    \caption{Slot description examples.}
    \label{fig:slotdesc}
\end{figure}





\section{Related Work}
\paragraph{Dialogue State Tracking} has been of broad interest to the dialogue research community~\cite{williams2007partially,williams2014dialog,heck2020trippy,liu2020attention,wu2020tod,madotto2020language}. Current state-of-the-art models~\cite{Chen2020SchemaGuidedMD,lin2020mintl,heck2020trippy,hosseini2020simple,ye2021slot,li2020coco} trained with extensive annotated data have been shown promising performance in complex multi-domain conversations~\cite{budzianowski2018multiwoz}. However, collecting large amounts of data for every domain is costly and inefficient. To address this issue, several methods~\cite{wu2019transferable,zhou2019multi} have proposed for transferring prior knowledge of existing domains to new ones. On the other hand, \citet{campagna2020zero} proposed an abstract dialogue model that leverages the ontology and in-domain templates to generate a large amount of synthesized data for domain adaptation. Different from their method, in this paper, we utilize a pre-trained seq2seq model and slot descriptions for cross-domain DST without any in-domain data.

\paragraph{Slot Description} has been shown to be a promising technique in cross domain semantic parsing~\cite{bapna2017towards,shah2019robust,namazifar2020language}. To encourage this line of research in DST as well, MultiWOZ2.1~\cite{eric2019multiwoz} provides a further annotation for slot descriptions. \citet{rastogi2020towards} incorporated slot descriptions for facilitating cross domain DST, while \citet{gao2019dialog,gao2020machine} formulated DST as a question answering problem by casting a slot name into questions. However, these works did not show the effectiveness of slot descriptions, by comparing the performance of models with and without them. There is no study on how to construct slot descriptions. In this paper, we aim to fill this research gap by providing an empirical study on the different slot description formulations.

\begin{table}[]
\resizebox{0.48\textwidth}{!}{
\begin{tabular}{@{}ll@{}}
\toprule
Slot Type & \multicolumn{1}{c}{Slot Name}                                                                                                          \\ \midrule
Number    & \begin{tabular}[c]{@{}l@{}}hotel-book stay,  hotel-book people, hotel-stars, \\ train-book people, restaurant-book people\end{tabular} \\ \midrule
Location  & \begin{tabular}[c]{@{}l@{}}train-destination, train-departure,  taxi-destination,\\ taxi-departure\end{tabular}                        \\ \midrule
Time      & \begin{tabular}[c]{@{}l@{}}train-arriveby, train-leaveat, taxi-leaveat, \\ restaurant-book time, taxi-arriveby\end{tabular}            \\ \midrule
Boolean   & hotel-parking, hotel-internet                                                                                                          \\ \midrule
Name      & attraction-name, restaurant-name, hotel-name                                                                                           \\ \midrule
Day       & hotel-book day, train-day, restaurant-book day                                                                                         \\ \bottomrule
\end{tabular}
}
\caption{Slot type of slots in MultiWOZ. The full table is reported in Appendix A.1.}
\label{table:slottype}
\end{table}

\begin{table*}[!t]
\centering
\resizebox{0.9\textwidth}{!}{
\begin{tabular}{@{}lcccccc@{}}
\toprule
\multirow{2}{*}{\textbf{Model}} & \multicolumn{6}{c}{\textbf{Joint Goal Accuracy}}                                                               \\
                       & Attraction         & Hotel      & Restaurant          & Taxi                & Train      & \textbf{Average}    \\ \midrule
TRADE                  & 19.87               & 13.70       & 11.52                & 60.58                & 22.37       & 25.76      \\
SUMBT*                  & 22.60               & \textbf{19.80}       & 16.50                & 59.50                & 22.50       & 28.18      \\
SimpleTOD++            & 28.01±1.30          & 17.69±1.00    & 15.57±1.54          & 59.22±0.95          & 27.75±1.16 & 29.65±0.58 \\
T5DST                  & \textbf{32.66}±0.10          & 18.73±1.67 & \textbf{20.55}±0.96          & \textbf{64.62}±0.24          & \textbf{31.27}±0.47 & \textbf{33.56}±0.54 \\ \midrule
w/ Human               & 31.92±1.42         & 20.72±0.35 & 20.09±0.67          & 64.12±0.28          & 28.83±1.28 & 33.14±0.17 \\
w/ Naive               & 32.98±0.60          & 20.23±1.11 & 20.01±2.91          & 63.59±0.23          & 30.04±4.31 & 33.37±1.36 \\
w/ Slot Value          & 32.86±0.56         & 20.03±0.87 & 16.65±0.37          & \textbf{65.09}±0.12 & 29.66±2.75 & 32.86±0.48 \\
w/ Question            & 32.45±0.39         & 19.79±1.18 & \textbf{21.82}±0.91 & 64.40±0.27           & 32.61±1.38 & 34.21±0.63 \\
w/ Slot Type           & \textbf{33.09}±1.60 & \textbf{21.21}±0.61 & 21.65±1.07          & 64.62±0.55          & \textbf{35.42}±1.42 & \textbf{35.20}±0.59  \\ \bottomrule
\end{tabular}
}
\caption{Zero-shot cross-domain results in MultiWOZ 2.0. We run each experiment three times with different random seeds, and report the mean and standard deviation. Note that the reported averaged zero shot joint goal accuracy is not comparable to multi-domains joint goal accuracy. *Result from~\cite{campagna2020zero}.}
\label{table:zeroshot}
\end{table*}

\section{Methodology}
\subsection{T5DST}
The design of our model follows the basis of generative question answering models. As illustrated in Figure~\ref{fig:diagram}, given a dialogue history which consists of an alternating set of utterances from two speakers, denoted as $\mathcal{C}_t=\{U_1,R_1, \dots, R_{t-1},U_t\}$, we add the \textit{"user:"} and \textit{"system:"} prefixes to the user and system utterance respectively. Then all the utterances and slot names $s_i$ are concatenated into a single sequence, i.e., \textit{user:$U_1$ $\dots$system:$R_{t-1}$ user:$U_{t}$ [sep] $s_i$}. The sequence is used as the input to the encoder, and the decoder generates the corresponding slot value $v_i$:

\begin{equation}
v_i= Seq2seq(\mathcal{C}_t, s_i).
\end{equation}
The learning objective of this generation process is minimizing the negative log-likelihood of $v_i$ given $\mathcal{C}_t$ and $s_i$, that is,
\begin{equation}
\mathcal{L} = -\sum_{i}^{n} \log p(v_i | \mathcal{C}_t, s_i),
\end{equation}
where $n$ is the number of slots to be tracked.

We initialize the model parameters with T5~\cite{raffel2020exploring}, an encoder-decoder Transformer with relative position embeddings~\cite{shaw2018self} pre-trained on a massive amount of English text. We denote our model as \textit{T5DST}. To incorporate slot descriptions into \textit{T5DST}, we replace the slot name with its corresponding slot description as the model input.

\subsection{Slot Type Informed Descriptions}
Although different slots may have distinguishing names, they can share the same slot type. As shown in Table 1, the slot type of \textit{hotel-stars} and \textit{restaurant-book people} are both number slots, while \textit{hotel-internet} and \textit{hotel-parking} are both boolean slots. In light of these observations, we hypothesize that adding slot type information to the slot description facilitates the knowledge transfer among different slots. We construct a template for each slot type that follows \textit{"[slot type] of [slot] of the [domain]"}. We denote such a slot description as \textit{Slot Type}. More details are available in Appendix A.1.



\begin{table*}[!t]
\centering
\resizebox{\textwidth}{!}{
\begin{tabular}{@{}lccccccccccccccc@{}}
\toprule
\multirow{2}{*}{\textbf{Model}} & \multicolumn{3}{c}{Attraction}                  & \multicolumn{3}{c}{Hotel}                        & \multicolumn{3}{c}{Restaurant}                   & \multicolumn{3}{c}{Taxi}                         & \multicolumn{3}{c}{Train}                        \\
                                & 1\%            & 5\%            & 10\%          & 1\%            & 5\%            & 10\%           & 1\%            & 5\%            & 10\%           & 1\%            & 5\%            & 10\%           & 1\%            & 5\%            & 10\%           \\ \midrule
TRADE                           & 35.88          & 57.55          & 63.12         & 19.73          & 37.45          & 41.42          & 42.42          & 55.70           & 60.94          & 63.81          & 66.58          & 70.19          & 59.83          & 69.27          & 71.11          \\
DSTQA                           & N/A            & \textbf{70.47} & \textbf{71.60} & N/A            & 50.18          & 53.68          & N/A            & 58.95          & \textbf{64.51} & N/A            & 70.90           & 74.19          & N/A            & 70.35          & 74.50           \\
T5DST w/ Slot Type              & \textbf{58.77} & 65.72          & 69.54         & \textbf{43.07} & \textbf{50.71} & \textbf{54.86} & \textbf{57.63} & \textbf{61.86} & 63.47          & \textbf{70.12} & \textbf{73.67} & \textbf{74.70} & \textbf{70.82} & \textbf{74.18} & \textbf{77.57} \\ \bottomrule
\end{tabular}
}
\caption{Few-shot experimental results in MultiWOZ 2.0. We evaluate our proposed model with 1\%, 5\%, and 10\% in-domain data, against TRADE~\cite{wu2019transferable} and DSTQA~\cite{zhou2019multi}. }
\label{table:fewshot}
\end{table*}

\section{Experiments}
\subsection{Dataset and Evaluation}
We evaluate the proposed method on the MultiWOZ 2.0 dataset~\cite{budzianowski2018multiwoz}, which has 7 domains.
We use the pre-processing and evaluation setup from \citet{wu2019transferable}, where restaurant, train, attraction, hotel, and taxi domains are used for training, as the test set only contains these 5 domains. 

In the zero-shot cross-domain experiments, the models are first trained with four domains and then evaluated on the test-set of the unseen domain. Joint goal accuracy is used to evaluate the performance of the models. The generated dialogue states are considered to be correct if and only if all of the predicted values exactly match the oracle values.


\subsection{Implementation}
We implement T5DST\footnote{Source code is available in \url{https://github.com/facebookresearch/Zero-Shot-DST}} based on the T5-small (60M parameters) model which has 6 encoder-decoder layers and the hidden size $d_{model}=512$. All models are trained using an AdamW~\cite{loshchilov2018decoupled} optimizer with the initial learning rate of $0.0001$. In all cross-domain zero-shot experiments, we train the models with batch size 128 for 5 epochs. For the few-shot experiments, the models
are first trained on 4 domains for 5 epochs then fine-tuned with 
1\%, 5\% and 10\% of target domain data for 10 epochs. For full shot training, we train our model for at most 10 epochs with batch size 64 and early stop according to the loss in the validation set. Other hyper-prameters are same as zero-shot cross-domain setting. We use 8 NVIDIA V100 GPUs for all of our experiments. We use greedy decoding in test time.


\subsection{Baselines}
\subsubsection{Models}
\paragraph{TRADE.} Transferable dialogue state generator~\cite{wu2019transferable} which utilizes copy mechanism to facilitate domain knowledge transfer.

\paragraph{SUMBT.} Slot-utterance matching belief tracker~\cite{lee2019sumbt} based on the language model BERT~\cite{devlin2018bert}.


\paragraph{DSTQA.} Dialogue state tracking via question answering\footnote{We are aware of STARC~\cite{gao2020machine}. However, we are not able to compare to our results with their results because they use different training data.} over ontology graph~\cite{zhou2019multi}.

\paragraph{SimpleTOD++.} SimpleTOD~\cite{hosseini2020simple} uses a single causal language model GPT2~\cite{radford2019language} to generate the dialogue states. To adapt this model to a zero-shot cross-domain setting, we also provide the slot name as the model input. We denote this model as SimpleTOD++.

\subsubsection{Slot Description Variants}

\paragraph{Human.} Human annotated slot descriptions collected in MultiWOZ2.1~\cite{eric2019multiwoz} and used in MultiWOZ2.2~\cite{zang2020multiwoz}.

\paragraph{Naive.}  Simple transformation of the slot name from \textit{"domain-slot"} to \textit{"[slot] of the [domain]"}.

\paragraph{Slot Value.} Following recent works~\cite{zhang2019find,rastogi2020towards}, slots are divided into \textit{categorical} and \textit{non-categorical} slots. For categorical slots, we incorporate the candidate values into the slot description, i.e., \textit{"[slot] of the [domain] is [value-1] or [value-2]?"}. The order of values is random. For non-categorical slots, their descriptions are the same as aforementioned \textit{Naive}.

\paragraph{Question.} Similar to \cite{gao2019dialog,gao2020machine}, we reformulate the slot into a natural language question, i.e., \textit{"What is the [slot] of the [domain] that is the user interested in?"}.

\subsection{Results \& Discussion}
\subsubsection{Zero-Shot Cross-Domain}

The results of the zero-shot cross domain experiments are shown in Table~\ref{table:zeroshot}. Overall, T5DST achieves significantly higher performance in terms of averaged joint goal accuracy compared to the three baseline models TRADE, SUMBT, and SimpleTOD++. These results demonstrate that our model can effectively capture the slot-context relation, and thus generalize better in unseen domains.

\begin{figure*}[!ht]
    \centering
    \includegraphics[width=\linewidth]{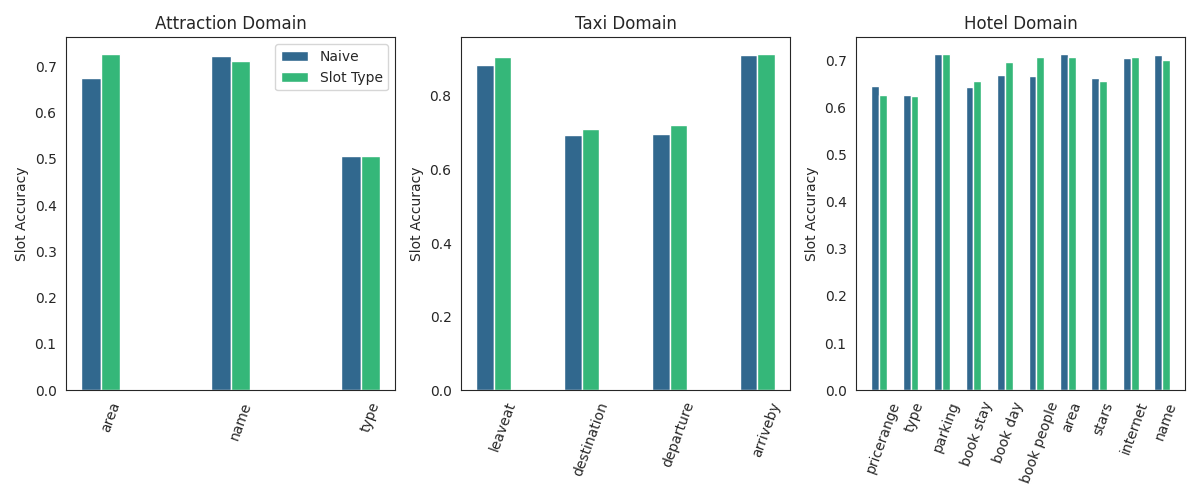}
    \caption{Slot accuracy in attraction, taxi, and hotel domains of MultiWOZ 2.0.}
    \label{fig:slotacc1}
\end{figure*}

\begin{figure}[!ht]
    \centering
    \includegraphics[width=\linewidth]{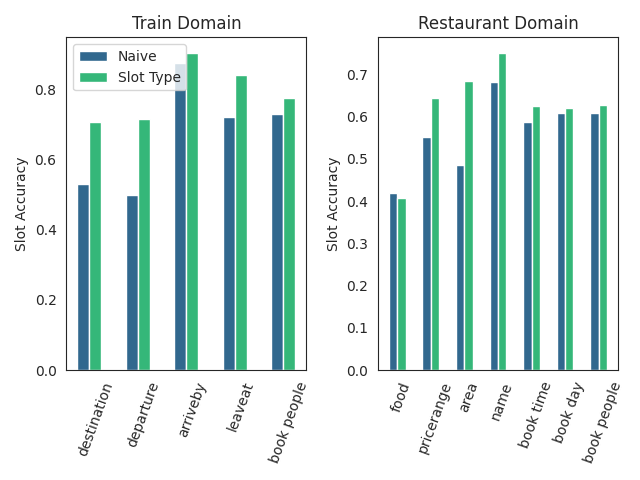}
    \caption{Slot accuracy in train and restaurant domains of MultiWOZ 2.0.}
    \label{fig:slotacc2}
\end{figure}

Replacing slot-names with human annotated slot descriptions does not bring improvement to the zero-shot performance. This might because of the diverse and inconsistent human descriptions among different domains. For example, the human descriptions of \textit{attraction-area} and \textit{restaurant-area} are \textit{"area to search for attractions"} and \textit{"area or place of the restaurant"} respectively. Such inconsistent descriptions increase the challenge on slot understanding in the zero-shot learning setting. the model using naive slot descriptions gives similar performance to the one that uses original slot names. The two approaches lead to similar semantic representation of the slots. In contrast, incorporating slot values hurts the learning, leading to a lower joint goal accuracy in the restaurant domain. We observe that even though adding value candidates improve some of the categorical slots (e.g., restaurant-area 68.35\% $\rightarrow$ 82.25\% slot accuracy), it hurts the unseen non-categorical slots (e.g., restaurant-food 40.63\% $\rightarrow$ 26.10\% slot accuracy). These non-categorical slots are usually the bottlenecks of joint goal accuracy. Finally, models trained with question style descriptions improves the performance in some domains, but fails in the others.

Our proposed slot type informed descriptions consistently improves the zero-shot performance of T5DST in all the domains. It produced an average of 2\% joint goal accuracy improvement compared to human labeled and naive description formulations. This result indicates that slot type information may better capture the shared property (e.g., time, location) among different slots, thus facilitating the domain knowledge transferring for DST.

Figure \ref{fig:slotacc1} and \ref{fig:slotacc2} show the slot accuracy of models using \textit{Naive} and \textit{Slot Type} description. Compared to naive description, we obverse significant gain of time slots (e.g., arrive by and leave at), location slots (e.g., departure and destination), and number slots (e.g., book stay and book people) by adding slot type information. We conjecture that explicit information about the target value (i.e., slot type) is important in the low resource condition when the model does not have enough data to capture the semantic meaning of a new slot.

\subsubsection{Few-Shot Cross-Domain}
We further conduct experiments in few-shot cross-domain settings, as in ~\cite{wu2019transferable,zhou2019multi}, where the models are first trained on 4 domains then fine-tuned with 1\%, 5\% and 10\% of target domain data. As shown in Table \ref{table:fewshot}, our model outperforms the DSTQA model in 4 out of 5 domains. Moreover, our approach is more practical in a real-world learning scenario as it does not require the supervision of a full ontology graph. We also conduct the full shot experiments and compare our model with previous methods. The reults are reported in Appendix A.2.



\section{Conclusion}
In this paper, we propose leveraging large scale pre-trained models with an effective slot description formulation to tackle the zero-shot cross-domain DST challenge. Specifically, we propose T5DST, a novel generative DST model based on the T5 language model, and incorporate Slot Type Informed Descriptions to facilitate cross-domain knowledge transfer. In the evaluation on the MultiWOZ dataset, our approach substantially improves existing results in both the zero-shot and few-shot settings. 





\bibliography{anthology,custom}
\bibliographystyle{acl_natbib}

\clearpage
\appendix

\section{Appendices}

\label{sec:appendix}
\subsection{Slot Type Informed Description Construction}
As shown in Table \ref{table:slottype_full}, each slot type has one prefix for appending to the beginning of the description. We used three different templates to construct the slot description. For all the booking slots (e.g., book people), we use \textit{"[prefix] [slot] for the [domain] booking"}. For boolean slots, we use \textit{"[prefix] [slot] in the [domain]"}. And for all the others, we use \textit{"[prefix] [slot] of the [domain]"}. When a slot name (e.g., \textit{train-day}) overlap with the slot type (e.g., \textit{day}) or a slot does not fall into any slot type category (others), we simply set the prefix as an empty string.



\subsection{Full Shot Results}
To understand the full shot performance of our T5DST model and whether slot description is still helpful when there is enough training data, we also conduct the experiments in a full data setting. As shown in Table \ref{table:fullshot}, using slot description only improves the joint goal accuracy by 0.56\% in MultiWoz 2.0 and 0.30\% in MultiWoz 2.1, which indicates that the description is less effective when there is a large amount of data for training. 

Compared to prior models with zero-shot capability, T5DST shows promising performance. Compared to other state-of-the-art models that optimized for full shot training, our model achieve competitive results in MultiWoz 2.0, but inferior results on MultiWoz 2.1. We notice that there are many training strategies (e.g., token masking~\cite{kim2019efficient,heck2020trippy}), additional supervision~(e.g., full ontology~\cite{Chen2020SchemaGuidedMD}), and label cleaning strategies~\cite{heck2020trippy}) that may impact final full-shot result. We also expect higher performance with a larger T5 model, such as T5-base or T5-large. However, achieving SOTA in full-scale training is out of the scope of this work.

\begin{table*}[t]
\resizebox{1\textwidth}{!}{
\begin{tabular}{@{}llll@{}}
\toprule
Slot Type & \multicolumn{1}{c}{Slot Name}                                                                                                                                & \multicolumn{1}{c}{Prefix} & Examples                               \\ \midrule
Number    & \begin{tabular}[c]{@{}l@{}}hotel-book stay,  hotel-book people, hotel-stars, \\ train-book people, restaurant-book people\end{tabular}                       & number of                  & number of people for the hotel booking \\ \midrule
Location  & \begin{tabular}[c]{@{}l@{}}train-destination, train-departure,  taxi-destination,\\ taxi-departure\end{tabular}                                              & location of                & location of destination of the train   \\ \midrule
Time      & \begin{tabular}[c]{@{}l@{}}train-arriveby, train-leaveat, taxi-leaveat, \\ restaurant-book time, taxi-arriveby\end{tabular}                                  & time of                    & time of arrive by of the train         \\ \midrule
Boolean   & hotel-parking, hotel-internet                                                                                                                                & whether have               & whether have parking in the hotel      \\ \midrule
Name      & attraction-name, restaurant-name, hotel-name                                                                                                                 & -                          & name of attraction                     \\ \midrule
Day       & hotel-book day, train-day, restaurant-book day                                                                                                               & -                          & day for the hotel booking              \\ \midrule
Others    & \begin{tabular}[c]{@{}l@{}}hotel-type, attraction-type, hotel-area, attraction-area, \\ restaurant-food, restaurant-pricerange, restaurant-area\end{tabular} & -                          & type of the hotel                      \\ \bottomrule
\end{tabular}
}
\caption{\textit{Slot Type} description examples. We define one prefix for each slot type. The prefix is empty when a slot name overlap with the slot type or a slot does not fall into any slot type category (others).}
\label{table:slottype_full}
\end{table*}

\begin{table*}[!ht]
\centering
\resizebox{0.95\textwidth}{!}{
\begin{tabular}{@{}lcccc@{}}
\toprule
\textbf{}         & \textbf{}            & \textbf{}                    & \multicolumn{2}{c}{\textbf{Joint Goal Accuracy}} \\ \cmidrule(l){4-5} 
\textbf{Models}   & \textbf{\#Parameter} & \textbf{Zero-shot Inference} & \textbf{MWoz 2.0}       & \textbf{MWoz 2.1}      \\ \midrule
TRADE~\cite{wu2019transferable}             & -                    & \cmark                            & 48.62                   & 45.6                   \\
STARC~\cite{gao2020machine}             & 110M                 & \cmark                            & -                       & 49.48                  \\
SUMBT~\cite{lee2019sumbt}             &                      & \cmark                            & 49.06                   & -                      \\
SGD-baseline~\cite{rastogi2020towards}      & 110M                 & \cmark                            & -                       & 43.4                   \\
T5DST             & 60M                  & \cmark                            & 52.86                   & 51.91                  \\
T5DST + Slot Type & 60M                  & \cmark                            & 53.42                   & 52.21                  \\ \midrule

DSTQA w/o span~\cite{zhou2019multi}    & -                    & \xmark                            & 51.44                   & 51.17                  \\
MinTL (BART)~\cite{lin2020mintl}    & 400M                    & \xmark                            & 52.10                   & 53.67                  \\
SOM-DST~\cite{kim2019efficient}           & 340M                 & \xmark                            & 52.32                   & 53.68                  \\
SST~\cite{Chen2020SchemaGuidedMD}               & 110M                 & \xmark                            & 51.17                   & 55.23                  \\
TripPy~\cite{heck2020trippy}            & 110M                 & \xmark                            & -                       & 55.29                  \\
SimpleTOD~\cite{hosseini2020simple}         & 110M                 & \xmark                            & -                       & 55.76                  \\ \bottomrule
\end{tabular}
}
\caption{Full shot results on MultiWOZ 2.0 and MultiWOZ 2.1.}

\label{table:fullshot}
\end{table*}



\end{document}